\documentclass{article}
\usepackage[utf8]{inputenc} % allow utf-8 input
\usepackage[T1]{fontenc}    % use 8-bit T1 fonts
\usepackage[a4paper, total={6in, 8in}]{geometry}
\usepackage{amsmath,amsthm}
\usepackage[hyphens]{url}
\usepackage{hyperref}       % hyperlinks
\hypersetup{colorlinks, pdfencoding=auto, psdextra}
\makeatletter
\newcommand\blx@noerroretextools\relax
\makeatother
\usepackage[style=authoryear-comp,mincitenames=1,maxcitenames=1,maxbibnames=9,backend=biber]{biblatex}
\usepackage{authblk}
\usepackage{booktabs}       % professional-quality tables
\usepackage{amsfonts}       % blackboard math symbols
\usepackage{nicefrac}       % compact symbols for 1/2, etc.
\usepackage{microtype}      % microtypography
\usepackage{xcolor}         % colors
\usepackage{parskip}
\usepackage{subcaption}

\usepackage{amsmath,amsthm}
\usepackage{makecell}
\usepackage{nicefrac}
\usepackage{enumitem}
\usepackage[ruled]{algorithm2e}
\setlist{nosep,leftmargin=*}

\usepackage{mdl}
\usepackage{autonum}

\AtEveryBibitem{
    \clearfield{month}
    \clearfield{day}
    \clearfield{urldate}
    \clearfield{urlyear}
    \clearfield{urlmonth}
    \clearfield{urlday}
    \clearfield{note}
    \clearfield{eventtitle}
    \clearfield{eventdate}
    \ifentrytype{online}{}{
        \clearfield{url}%
    }
    \ifentrytype{misc}{
        \clearfield{doi}%
    }{}
}
\DeclareFieldFormat[article,inbook,incollection,inproceedings,patent,thesis,unpublished]{citetitle}{#1\isdot}
\DeclareFieldFormat[article,inbook,incollection,inproceedings,patent,thesis,unpublished]{title}{#1\isdot}

\allowdisplaybreaks[4]
\title{Harnessing the Power of Vicinity-Informed Analysis for Classification under Covariate Shift}

\author[1,3]{Mitsuhiro Fujikawa\thanks{mitsuhiro@mdl.cs.tsukuba.ac.jp}}
\author[1,3]{Yohei Akimoto\thanks{akimoto@cs.tsukuba.ac.jp}}
\author[2,3]{Jun Sakuma\thanks{sakuma@c.titech.ac.jp}}
\author[1,3]{Kazuto Fukuchi\thanks{fukuchi@cs.tsukuba.ac.jp, Corresponding author}}
\affil[1]{University of Tsukuba, Japan}
\affil[2]{Tokyo Institute of Technology, Japan}
\affil[3]{RIKEN AIP, Japan}

\begin{document}

\maketitle

\begin{abstract}\noindent
    Transfer learning enhances prediction accuracy on a target distribution by leveraging data from a source distribution, demonstrating significant benefits in various applications. This paper introduces a novel dissimilarity measure that utilizes vicinity information, i.e., the local structure of data points, to analyze the excess error in classification under covariate shift, a transfer learning setting where marginal feature distributions differ but conditional label distributions remain the same. We characterize the excess error using the proposed measure and demonstrate faster or competitive convergence rates compared to previous techniques. Notably, our approach is effective in the support non-containment assumption, which often appears in real-world applications, holds. Our theoretical analysis bridges the gap between current theoretical findings and empirical observations in transfer learning, particularly in scenarios with significant differences between source and target distributions.
\end{abstract}

\section{Introduction}

Transfer learning is a technique for enhancing prediction accuracy by utilizing a sample from a distribution~(source distribution), which is different from the distribution where predictions are actually made~(target distribution). Existing empirical studies of transfer learning have shown significant accuracy improvements by leveraging a sample from the source distribution~\autocite{dai_boosting_2007,shi_actively_2008,donahue_decaf_2014,schreuder_classification_2021,ginsberg_learning_2022,zhang_adapting_2023,sui_unleashing_2023}. However, these findings are valid only for the datasets tested, leaving the effectiveness in unexplored scenarios uncertain. Theoretical analysis, on the other hand, offers broader assurances of these enhancements across various situations.

Our paper primarily focuses on theoretical analysis of classification under the covariate-shift~\autocite{shimodaira_improving_2000} environment. Covariate-shift refers to a scenario where, despite the relationships between features and labels remaining consistent across source and target distributions, the marginal distributions of features differ. A key property in characterizing the success of transfer learning under covariate-shift is consistency with respect to the source sample size. A classification algorithm is deemed consistent with respect to source sample size if its error rate decreases to the optimal one as the size of the source sample increases indefinitely, highlighting the achievability of the optimal classifier by utilizing the source sample. The main focus of our theoretical analyses is to validate source sample-size consistency of the constructed classification algorithm under the covariate-shift.

Several theoretical techniques have been developed to analyze classification error under the covariate-shift setup; however, most of them lack the capability to validate source sample size consistency. For example, many researchers have derived upper bounds on the generalization error using distance measures between source and target distributions~\autocite{ben-david_theory_2010,mansour_domain_2023,ruan_optimal_2021,aminian_information-theoretical_2022,park_calibrated_2020}. These techniques are applicable to a broad range of situations since they do not make assumptions about the source and target distributions. However, they might fail to validate source sample size consistency because the distance measures used in these techniques may remain positive even when the source sample size tends to infinity.

Only a few theoretical results can prove the achievability of source sample size consistency. One notable result is the work by \textcite{pathak_new_2022}, who deal with the nonparametric regression problem under covariate-shift and analyze the regression error using the following dissimilarity measure. Let $P$ and $Q$ be source and target distributions whose marginals for features are denoted as $P_X$ and $Q_X$, respectively. Let $\mathcal{X}$ be the universe of features equipped with a metric $\rho$. Given a level $r > 0$, their dissimilarity measure is defined~\footnote{We interpret as $\Delta_{\mathrm{PMW}}(P,Q;r) = \infty$ if $Q_X(P_X(B(X, r)) > 0) < 1$.} as
\begin{align}
    \Delta_{\mathrm{PMW}}(P,Q;r) = \int_{\mathcal{X}} \frac{1}{P_X(B(x,r))} Q_X(dx), \label{newsim}
\end{align}
where $B(x,r) = \left\{x' \in \mathcal{X} : \rho\left(x,x'\right) \leq r\right\}$ is the closed ball of radius $r$ centered at $x$. We use the notation $B(x,r)$ when $\rho$ is clear from the context. \textcite{pathak_new_2022} demonstrate that a consistent regression algorithm exists if the dissimilarity measure in \cref{newsim} is less than a polynomial order of $r^{-1}$. This result can be readily extended to the classification case by utilizing the results of \textcite{kpotufe_marginal_2021,galbraith_classification_2024}~(See \cref{sec:main}).

One significant limitation of their techniques is the inability to prove source sample-size consistency in situations where the support of the source distribution does not cotain that of the target distribution. In such situations, their dissimilarity measure in \cref{newsim} becomes infinite because the probability $P_X(B(x,r))$ becomes zero for small $r$ at points $x$ that appear in the target distribution but not in the source distribution. However, these situations are prevalent in real-world applications, and empirical evidence indicates the effectiveness of transfer learning even under a support non-containment environment. For instance, several researchers~\autocite{hoyer_mic_2023,zhou_unsupervised_2024,zhu_patch-mix_2023,westfechtel_combining_2023} have demonstrated the success of their methods on the Office-Home dataset~\autocite{venkateswara_deep_2017}, in which source and target datasets consist of images from different domains, including artistic depictions, clipart images, images without backgrounds, and real-world images. The appearances of images across different domains are considerably different, suggesting non-containment of supports. Consequently, the current theoretical framework fails to capture the success demonstrated in this example, revealing a gap between existing theoretical results and real-world observations. This discrepancy highlights the need for a theoretical approach that can account for the effectiveness of transfer learning in scenarios where the support containment assumption does not hold.

\paragraph{Our dissimilarity measure and contributions.}
This study bridges this gap by introducing a novel dissimilarity measure and characterizing the classification error under covariate-shift using the proposed measure. Our dissimilarity measure is defined~\footnote{We interpret as $\Delta_{\mathcal{V}}(P,Q;r) = \infty$ if $Q_X(\sup_{x'\in\mathcal{V}(X)} P_X(B(x',r)) > 0) < 1$. } as follows:
\begin{align}
    \Delta_{\mathcal{V}}(P,Q;r) = \int_{\mathcal{X}} \inf_{x' \in \mathcal{V}(x)} \frac{1}{P_X(B(x',r))} Q_X(dx), \label{oursim}
\end{align}
where $\mathcal{V}(x)$ denotes the set of the vicinity surrounding the point $x$, whose rigorous definition will be explored in \cref{sec:main}. The only difference between \cref{oursim} and \cref{newsim} is that \cref{oursim} takes the infimum over $\mathcal{V}(x)$ when evaluating the inverse probability, whereas \cref{newsim} evaluates the inverse probability at $x$. By taking the infimum, we may avoid evaluating the inverse probability at points where the probability $P_X(B(x,r))$ becomes zero. This makes the resultant dissimilarity value finite even when the support of the source distribution does not contain that of the target distribution.

The utility of our dissimilarity measure in \cref{oursim} is highlighted by the following contributions:
\begin{itemize}
    \item We derive an upper bound on the excess error under covariate-shift and provide a characterization of it via the dissimilarity measure in \cref{oursim}. A notable insight from this characterization is the existence of a classification algorithm that is consistent for the source sample size, which can validate the source sample-size consistency even under the support non-containment environment:
    \begin{theorem}[Informal]\label{thm:informal}
        Under certain conditions, there exists a classification algorithm that is consistent for the source sample size if $\Delta_{\mathcal{V}}(P,Q;r)$ is less than a polynomial order of $r^{-1}$.
    \end{theorem}
    \cref{thm:informal} provides the same characterization of the source sample size consistency as shown by \textcite{pathak_new_2022}, except it uses our dissimilarity measure $\Delta_{\mathcal{V}}$ instead of their measure $\Delta_{\mathrm{PMW}}$.
    \item We propose novel notions of $\Delta$-transfer-exponent and $\Delta$-self-exponent for a dissimilarity measure $\Delta$. These notions are a generalization of the concept of $\alpha$-families provided by \textcite{pathak_new_2022}. Our notions of the $\Delta$-transfer-exponent and $\Delta$-self-exponent universally characterize the upper bounds obtained by \textcite{pathak_new_2022}, \textcite{kpotufe_marginal_2021}, and our own work, thereby enabling a fair comparison among these upper bounds. Indeed, we prove that an upper bound on the excess error derived from our dissimilarity measure in \cref{oursim} always exhibits faster or competitive convergence rates compared to the rates of the upper bounds obtained from the existing measures provided by \textcite{pathak_new_2022} and \textcite{kpotufe_marginal_2021}. This improvement in convergence rates highlights the advantage of incorporating vicinity information in the dissimilarity measure.
    \item We conducted experiments comparing our method with \textcite{pathak_new_2022}'s approach on synthetic datasets with support non-containment setups. The results demonstrate the tightness of our derived upper bound and showcase our method's ability to achieve source sample-size consistency in the support non-containment setting, a feat unattained by the existing method.
\end{itemize}

All the missing proofs can be found in \cref{sec:missing-proofs}.

\section{Preliminaries}
\paragraph{Notations} For a probability measure $P$ and a positive integer $k$, let $P^k$ denote the $k$-fold product measure of $P$. Given a probability measure $P$ and a random variable $X$, we denote $\mathbb{E}_P[X]$ as the expectation of $X$ under the distribution $P$. For an event $\mathcal{E}$, we use $\mathbbm{1}\{\mathcal{E}\}$ to denote the indicator function. Given a metric space $(\mathcal{X},\rho)$ and a radius $r > 0$, let denote the closed sphere centered at $x \in \mathcal{X}$ with radius $r$ as $B(x,r)=\cbrace{x' \in \mathcal{X} : \rho(x,x') \le r}$.

\subsection{Classification under Covariate-shift}
Consider a classification problem under the covariate shift setup. Let $X$ be a random variable representing the input to a classifier, equipped with a compact metric space $(\mathcal{X},\rho)$ of diameter $D_{\mathcal{X}}$, and let $Y$ be a random variable signifying the binary label, i.e., with a universe of $\mathcal{Y}=\cbrace{0,1}$. The learner has access to a sample composed of labeled data from two distributions: the source distribution $P$ and the target distribution $Q$. The labeled data from the source and target distributions are denoted as $(\mathbf{X},\mathbf{Y})_P = \{(X_i,Y_i)\}^{n_P}_{i=1} \sim P^{n_P}$ and $(\mathbf{X},\mathbf{Y})_Q = \{(X_i,Y_i)\}^{n_P + n_Q}_{i=n_P + 1} \sim Q^{n_Q}$, respectively, where $n_P$ and $n_Q$ represent the source and target sample sizes. Given the sample $\paren{\mathbf{X},\mathbf{Y}} = \paren{\mathbf{X},\mathbf{Y}}_P \cup \paren{\mathbf{X},\mathbf{Y}}_Q$, the learner's objective is to construct a classifier $h : \mathcal{X} \rightarrow \mathcal{Y}$ that minimizes its error rate for the target distribution, defined as:
\begin{align}
    err_{Q}\paren*{h}=\mathbb{E}_{Q} \mathbbm{1}\cbrace*{h(X) \ne Y}.
\end{align}
For convenience, let $\mathcal{X}_P$ be the support of $P_X$, i.e., $\mathcal{X}_P=\cbrace{x\in \mathcal{X} : P_X\paren{B\paren{x,r}} > 0,\forall{r} > 0}$. Define $\mathcal{X}_Q$ similarly to $\mathcal{X}_P$.

Covariate-shift is a relationship between the source and target distributions, in which the marginal distributions of the input $X$ can differ between $P$ and $Q$, whereas the distributions of the label $Y$ conditioned on the input $X$ are identical. Let $P_X$ and $Q_X$ be the marginal source and target distributions of $X$, respectively. Let $P_{Y|X}$ and $Q_{Y|X}$ be the source and target distributions of $Y$ conditioned on $X$, respectively. Then, covariate shift is rigorously defined as follows:
\begin{definition}[Covariate-shift]
    The relationship between distribution $P$ and distribution $Q$ is covariate shift if there exists a measurable function $\eta : \mathcal{X} \rightarrow [0,1]$, called a regression function, such that $P_{Y|X}(Y=1|X) = Q_{Y|X}(Y=1|X) = \eta\left(X\right)$ $P_X$- and $Q_X$-almost surely.
\end{definition}
This definition indicates that, for example, traffic signs appearing in urban and rural areas may differ ($P_X \ne Q_X$), but their instructions are consistent regardless of the location ($P_{Y|X} = Q_{Y|X}$) in the context of sign recognition in an automated driving system.

\subsection{Excess Error}
The objective of our theoretical analyses is to elucidate the relationship between the source and target sample sizes~($n_P$ and $n_Q$) and the {\em excess error}. The excess error of a classifier $h$ is defined as the difference between the error of $h$ and the error incurred by the {\em Bayes classifier} $h^*$. The Bayes classifier, under the error metric $err_{Q}\left(h\right)$, is the classifier that minimizes $err_{Q}\left(h\right)$. The formal definition of the excess error is as follows:
\begin{definition}[excess error]
    The excess error of the classifier $h$ for the distribution $Q$ is given by:
    \begin{align}
      \mathcal{E}_Q\paren*{h}=err_Q\paren*{h}-err_Q\paren*{h^*}.
      \label{ee}
    \end{align}
\end{definition}
As the excess error approaches 0, the classifier $h$ approaches the performance of the ideal classifier. Under our setup, the Bayes classifier can be expressed as $h^*\left(x\right)=\mathbbm{1}\cbrace{\eta(x)\geq 1/2}$. Consequently, the set of points $x$ for which $\eta(x)=1/2$ can be considered as the correct decision boundary, as the Bayes classifier assigns the label 1 to points with $\eta(x) > 1/2$ and the label 0 to points with $\eta(x) < 1/2$.

\subsection{Difficulty in Classification under Distribution $Q$}
For the purpose of our analyses, we introduce the following common assumptions that stipulate the difficulty in classification under distribution $Q$.
\begin{definition}[Smoothness]\label{holder}
    A regression function $\eta$ is $(C_\alpha,\alpha)$-H\"older for $\alpha \in (0,1]$ and $C_\alpha > 0$ if $\forall x, x' \in \mathcal{X}, |\eta(x)-\eta(x')| \leq C_\alpha \cdot \rho(x,x')^\alpha$.
\end{definition}
\begin{definition}[Tsybakov's noise condition]\label{def:noise}
    A distribution $Q$ satisfies Tsybakov's noise condition with parameters $\beta > 0$ and $C_\beta > 0$ if $\forall t \geq 0, Q_X(0<|\eta(X)-\frac{1}{2}|\leq t) \leq C_\beta t^\beta$.
\end{definition}
The smoothness condition in \cref{holder} requires that the labels for similar inputs are likely to be the same. The noise parameters determine the probability of observing a label with a large amount of noise. It is worth noting that \textcite{kpotufe_marginal_2021,galbraith_classification_2024} conducted their analyses under the same assumptions. Similarly, \textcite{pathak_new_2022} employ assumptions regarding smoothness and noise; however, their assumptions differ slightly from ours, as they address a different problem: regression, while we focus on classification.

Our analyses will be conducted under the assumption that the target distribution satisfies both the smoothness and noise conditions.
\begin{definition}[$\mathrm{STN}(\alpha,\beta)$]
    A distribution $Q$ is $\mathrm{STN}(\alpha,\beta)$ if there exist some constants $C_\alpha>0$ and $C_\beta>0$ such that the regression function $\eta$ is $(C_\alpha,\alpha)$-H\"older, and $Q$ satisfies Tsybakov's noise condition with parameters $\beta$ and $C_\beta$.
\end{definition}

\section{Main Result}\label{sec:main}

Our main result is a characterization of the excess error under the covariate-shift setup using our dissimilarity measure in \cref{oursim}. Specifically, we characterize the excess error through the $\Delta$-transfer and $\Delta$-self exponents, which are quantities derived from the distributions $(P,Q)$ using the dissimilarity measure $\Delta$. We begin by introducing the definitions of the $\Delta$-transfer and $\Delta$-self exponents, followed by our characterization of the excess error. Furthermore, we reproduce the results of \textcite{pathak_new_2022} and \textcite{kpotufe_marginal_2021} using these exponents, enabling a fair comparison between our results and theirs. Through this comparison, we demonstrate that the excess error upper bound obtained using our dissimilarity measure consistently exhibits faster or competitive rates compared to existing methods.

\paragraph{Transfer and self exponents}
To characterize the excess error by some quantity of $(P,Q)$, we generalize the notion of the $\alpha$-family proposed by \textcite{pathak_new_2022}. Specifically, we characterize the excess error by the following quantities determined by a dissimilarity measure.
\begin{definition}[$\Delta$-transfer-exponent]\label{def:trans-expo}
    Given a dissimilarity measure $\Delta$, a distribution pair $\left(P,Q\right)$ has a $\Delta$-transfer-exponent of $\tau \in [0,\infty]$ if there exists a constant $C \ge 1$ such that
    \begin{align}
        \sup_{0 < r \leq D_{\mathcal{X}}} \paren*{r/D_{\mathcal{X}}}^{\tau} \Delta(P,Q;r) \leq C,
    \end{align}
    where $0 \cdot \Delta(P,Q;0) = 0$.
\end{definition}
\begin{definition}[$\Delta$-self-exponent]\label{def:self-expo}
    Given a dissimilarity measure $\Delta$, a distribution $Q$ has a $\Delta$-self-exponent of $\psi \in (0,\infty]$ if there exists a constant $C \ge 1$ such that
    \begin{align}
        \sup_{0 < r \leq D_{\mathcal{X}}} \paren*{r/D_{\mathcal{X}}}^{\psi} \Delta(Q,Q;r) \leq C.
    \end{align}
\end{definition}
\cref{def:trans-expo} and \cref{def:self-expo} imply that the dissimilarities $\Delta(P,Q;r)$ and $\Delta(Q,Q;r)$ decrease at a polynomial rate with respect to $r^{-1}$, with exponents $\tau$ and $\psi$, respectively. In other words, $\Delta(P,Q;r) = O(r^{-\tau})$ and $\Delta(Q,Q;r) = O(r^{-\psi})$ for a decreasing $r$. It is worth noting that our definitions of $\Delta$-transfer-exponent and $\Delta$-self-exponent are universal in the sense that we can exactly reproduce the quantities used in existing characterizations by choosing an appropriate dissimilarity measure $\Delta$, which will be discussed later.

\paragraph{Vicinity set}
Our characterization is based on $\Delta_{\mathcal{V}}$-transfer and $\Delta_{\mathcal{V}}$-self exponents, with an appropriate choice of the vicinity set function $\mathcal{V}(x)$. The choice of the vicinity set function $\mathcal{V}$ is formally given as follows:
\begin{align}
    \mathcal{V}(x) = \cbrace*{ x' \in \mathcal{X} : 2 C_\alpha \rho(x,x')^\alpha < \abs*{\eta(x) - \frac{1}{2} }} \cup \cbrace*{x}. 
    \label{eq:vicinity}
\end{align}
The vicinity set $\mathcal{V}(x)$ is the (nearly-)largest open ball centered at $x$ such that the labels of the Bayes classifier evaluated at points within the ball are consistent. We can expect that these vicinity points may share the same label information and thus are useful for predicting the label at $x$.

\paragraph{Our characterization}
As our characterization, we provide an upper bound on the excess error composed of the source and target sample sizes as well as transfer- and self-exponents. 
\begin{theorem} \label{thm:main}
    Given $\alpha \in (0,1]$, $\beta > 0$, and $\psi\in (0,\infty]$, suppose the target distribution $Q$ is $\mathrm{STN}(\alpha,\beta)$ and has $\Delta_{\mathcal{V}}$-self-exponent of $\psi$. Also, suppose $\left(P,Q\right)$ has $\Delta_{\mathcal{V}}$-transfer-exponent of $\tau$ for some $\tau \in (0,\infty]$. Then, there exists a classification algorithm which produces a classifier $\hat{h}$ such that for all $n_P > 0$ and $n_Q > 0$, 
    \begin{align}
        \mathbb{E}\bracket*{\mathcal{E}_Q(\hat{h})} \le 
         C\begin{dcases}
             \log\paren*{n_P+n_Q}\paren*{n_p^{\frac{1+\beta}{2 + \beta + \max\cbrace{1,\tau/\alpha}}} + n_Q^{\frac{1+\beta}{2 + \beta + \max\cbrace{1,\psi/\alpha}}}}^{-1}  & \textif \alpha = \tau \textor \alpha = \psi, \\
             \paren*{n_p^{\frac{1+\beta}{2 + \beta + \max\cbrace{1,\tau/\alpha}}} + n_Q^{\frac{1+\beta}{2 + \beta + \max\cbrace{1,\psi/\alpha}})}}^{-1}  & \otherwise, 
         \end{dcases}\label{the:mub2}
    \end{align}
    where $C > 0$ is some constant independent of $n_P$ and $n_Q$.
\end{theorem}
The implications of \cref{thm:main} are as follows:
\begin{enumerate}
    \item \cref{thm:main} directly establishes that the sufficient condition for the existence of a source sample size consistent classification algorithm is $\tau < \infty$. In this case, the exponent of $n_P$ is non-zero, indicating the algorithm's consistency with respect to the source sample size.
    \item In the non-transfer setting, the excess error decreases as the sample size increases, with an exponent of $-\frac{1+\beta}{2 + \beta + d/\alpha}$ for $d$-dimensional input, i.e., $\mathcal{X} \subset \mathbb{R}^d$~\autocite{audibert_fast_2007}. Our bound exhibits the same characterization, except that the dimensionality $d$ is replaced by the $\Delta_{\mathcal{V}}$-transfer- or $\Delta_{\mathcal{V}}$-self-exponent, corresponding to $n_P$ or $n_Q$, respectively. Indeed, the $\Delta_{\mathcal{V}}$-self-exponent plays a role similar to the dimensionality $d$, as it is smaller than $d$ for $\mathcal{X} \subset \mathbb{R}^d$~\footnote{This discussion is valid only when $\mathcal{X}$ is bounded. Exploring the unbounded case is one of our future directions.}.
    \item The $\Delta_{\mathcal{V}}$-transfer- and $\Delta_{\mathcal{V}}$-self-exponents characterize the dependency of the excess error on the source and target sample sizes, respectively. Indeed, the convergence rate of the excess error for $n_P$~(resp., $n_Q$) becomes faster as the $\Delta_{\mathcal{V}}$-transfer-exponent~(resp., $\Delta_{\mathcal{V}}$-self-exponent) decreases.
\end{enumerate}

\paragraph{Comparisons with \textcite{pathak_new_2022} and \textcite{kpotufe_marginal_2021}.}
We explore the comparison with the excess error upper bounds shown by \textcite{pathak_new_2022} and \textcite{kpotufe_marginal_2021}. As mentioned in the introduction, \textcite{pathak_new_2022} provide a characterization of the excess error through $\Delta_{\mathrm{PMW}}$ in \cref{newsim}. We can reproduce the results of \textcite{kpotufe_marginal_2021} via the transfer- and self-exponents of the following measures:
\begin{align}
    \Delta_{\mathrm{DM}}(Q,Q;r) =& \sup_{x \in \mathcal{X}_Q}\frac{1}{Q_X(B(x,r))}, \\
    \Delta_{\mathrm{BCN}}(Q,Q;r) =& \mathcal{N}(\mathcal{X}_Q, \rho, r), \\
    \Delta_{\mathrm{KM}}(P,Q;r) =& \sup_{x \in \mathcal{X}_Q}\frac{Q_X(B(x,r))}{P_X(B(x,r))},
\end{align}
where $\mathcal{N}(\mathcal{X}_Q, \rho, r)$ denotes the $r$-covering number of the set $\mathcal{X}_Q$. Building upon the measures $\Delta_{\mathrm{PMW}}$, $\Delta_{\mathrm{DM}}$, $\Delta_{\mathrm{BCN}}$, and $\Delta_{\mathrm{KM}}$, their upper bounds are reproduced as follows:
\begin{proposition}[\textcite{pathak_new_2022,kpotufe_marginal_2021}]\label{prop:exist-bounds}
    Given $\alpha \in (0,1]$ and $\beta > 0$, suppose the target distribution $Q$ is $\mathrm{STN}(\alpha,\beta)$. For $\psi\in (0,\infty]$ and $\tau \in (0,\infty]$, we suppose that the one of the following conditions holds:
    \begin{enumerate}
        \item $Q$ has the $\Delta_{\mathrm{PMW}}$-self-exponent of $\psi$, and $(P,Q)$ has $\Delta_{\mathrm{PMW}}$-transfer-exponent of $\tau$.
        \item $Q$ has the $\Delta_{\mathrm{DM}}$- or $\Delta_{\mathrm{BCN}}$-self-exponent of $\psi$, $(P,Q)$ has $\Delta_{\mathrm{KM}}$-transfer-exponent of $\tau-\psi$, and $\tau \ge \psi$.
    \end{enumerate}
    Then, there exists an algorithm that exhibits the excess error upper bound obtained by \cref{thm:main}. 
\end{proposition}
\cref{prop:exist-bounds} indicates that our bound in \cref{thm:main} coincides with theirs, except for using the self- and transfer-exponents with their measures.

Next, we compare the self- and transfer-exponents between our and their measures. 
\begin{proposition}\label{prop:rel-exps}
    For any pair of distributions $(P,Q)$, we have
    \begin{alignat}{3}
        \tau_{\Delta_{\mathcal{V}}} \le& \tau_{\Delta_{\mathrm{PMW}}} &\le& \tau_{\Delta_{\mathrm{KM}}} + &\min\{ \psi_{\Delta_{\mathrm{DM}}}, \psi_{\Delta_{\mathrm{BCN}}}\}, \\
        \psi_{\Delta_{\mathcal{V}}} \le& \psi_{\Delta_{\mathrm{PMW}}} &\le& &\min\{ \psi_{\Delta_{\mathrm{DM}}}, \psi_{\Delta_{\mathrm{BCN}}}\},
    \end{alignat}
    where $\tau_\Delta$ and $\psi_\Delta$ denotes the minimum $\Delta$-transfer- and $\Delta$-self-exponents $(P,Q)$ has.
\end{proposition}
\cref{prop:rel-exps} showcases that $\Delta_{\mathcal{V}}$ achieves the smallest transfer- and self-exponents, indicating that our measure can provide faster rates than those obtained by the existing measures.

\paragraph{Example}
We demonstrate that, unlike existing measures, our dissimilarity measure can validate the source sample size consistency even when $\mathcal{X}_Q \not\subseteq \mathcal{X}_P$. We provide a concrete example of $P$ and $Q$ to illustrate this property. Consider the case where $\mathcal{X} = \mathbb{R}$. Suppose $P_X$ and $Q_X$ are uniform distributions over $[-\frac{7}{8}, \frac{7}{8}]$ and $[-1,1]$, respectively, and the regression function is $\eta\left(x\right) = \frac{1}{2}x+\frac{1}{2}$. It is clear that $\mathcal{X}_P \subset \mathcal{X}_Q$, and hence $\mathcal{X}_Q \not\subseteq \mathcal{X}_P$. The self-exponents are equivalent, i.e., $\psi_{\Delta_{\mathcal{V}}} = \psi_{\Delta_{\mathrm{PMW}}} = \psi_{\Delta_{\mathrm{DM}}} = \psi_{\Delta_{\mathrm{BCN}}} = 1$. However, the probability $P_X(B(x,r))$ takes zero at $x \in [-1,-\frac{7}{8}-r)\cup(\frac{7}{8}+r,1]$ for a small $r$, causing the existing transfer-exponents to become infinite, i.e., $\tau_{\Delta_{\mathrm{PMW}}} = \tau_{\Delta_{\mathrm{KM}}} = \infty$. In contrast, our measure satisfies $\tau_{\Delta_{\mathcal{V}}} = 1$ because $\mathcal{V}(x) \cap [-\frac{7}{8}, \frac{7}{8}]$ is non-empty for any $x \in [-1,1]$, and the probability $P_X(B(x,r))$ is non-zero for any $x \in [-\frac{7}{8}, \frac{7}{8}]$. Consequently, our bound exhibits the rate of $\ln(n_P+n_Q)(n_P^{1/2} + n_Q^{1/2})^{-1}$, as $\alpha = 1$ and $\beta = 1$ in this case, achieving the source sample size consistency.

\section{Analyses}\label{sec:analyses}
To prove \cref{thm:main}, we provide an upper bound on the excess error of a specific classification algorithm, the $k$-nearest neighbor ($k$-NN) classifier proposed by \textcite{kpotufe_marginal_2021}. Given a point $X$ distributed according to $Q_X$ for which the label will be predicted, the $k$-NN classifier first estimates the regression function's output at $X$, denoted as $\eta(X)$, by computing the average of labels over the $k$ nearest neighbor points in $\paren{\mathbf{X},\mathbf{Y}}$. The predicted label is then determined to be 1 if the estimated value of $\eta(X)$ is greater than $\nicefrac{1}{2}$ and 0 otherwise. Formally, let $(X_{(1)},Y_{(1)}), ..., (X_{(k)}, Y_{(k)})$ be the $k$ nearest neighbors of $X$ and their corresponding labels. The estimated regression function is given by $\hat\eta_k(X) = \frac{1}{k}\sum_{i=1}^kY_{(i)}$, and the predicted label $\hat{h}_k(X)$ is determined as $\hat{h}_k(X) = \mathbbm{1}\cbrace{\hat\eta_k(X) \ge \nicefrac{1}{2}}$.

The goal of this section is to demonstrate that the upper bound shown in \cref{thm:main} is achievable by the $k$-NN classifier with an appropriate choice of $k$.
\begin{theorem}\label{thm:analyses-result}
    Under the same assumptions as in \cref{thm:main}, the $k$-NN classifier with 
    \begin{align}
        k = \floor*{\paren*{n_P^{\frac{1}{2 + \beta + \max\cbrace{1,\nicefrac{\tau}{\alpha}}}} + n_Q^{\frac{1}{2 + \beta + \max\cbrace{1,\nicefrac{\psi}{\alpha}}}}}^2},    
    \end{align}
    achieves the excess error upper bound shown in \cref{thm:main}.
\end{theorem}

The main challenge in proving \cref{thm:analyses-result} is linking the excess error of the $k$-NN classifier to the minimum inverse probability that appears in our dissimilarity measure in \cref{oursim}. To achieve this, we derive an upper bound on the excess error of the $k$-NN classifier using the {\em vicinity distance}, defined as, for $z, x \in \dom{X}$,
\begin{align}
    \rho_{\dom{V}}(z,x) = \inf_{x' \in \dom{V}(x)}\rho(z, x'). \label{eq:rho-v}
\end{align}
The vicinity distance $\rho_{\dom{V}}$ characterizes the minimum inverse probability, as it can be rewritten as 
\begin{align}
    \inf_{x' \in \mathcal{V}(x)}P_X^{-1}(B(x', r)) = P_X^{-1}\paren*{\inf_{x' \in \mathcal{V}(x)}\rho(X, x') \le r} = P_X^{-1}\paren*{\rho_{\dom{V}}(X, x) \le r}.
\end{align}
Therefore, characterizing the excess error using $\rho_{\dom{V}}$ is crucial for revealing its connection to our dissimilarity measure $\Delta_{\mathcal{V}}$. The details of this analysis will be explored in the subsequent subsection.

\subsection{Bounding Excess Error by Vicinity Distance}\label{sec:min-dist-bias}
This subsection aims to derive an upper bound on the excess error of the $k$-NN classifier using the vicinity distance $\rho_{\mathcal{V}}$. We first employ two existing techniques from \textcite{kpotufe_marginal_2021}: an upper bound on the excess error by the approximation error of $\hat\eta_k$ and the concept of implicit $1$-NNs. Then, we show that the approximation error of $\hat\eta_k$ is bounded above by the expected vicinity distance between an implicit $1$-NN and a point to be predicted.

\paragraph{Bounding via the approximation error of $\hat\eta_k$.}
We construct an upper bound on the excess error of the $k$-NN classifier using the approximation error of the estimated regression function $\hat\eta_k$. Define $g(X) = \abs{\eta(X) - \frac{1}{2}}$. For a random variable $Z$ (possibly) depending on $X$ and $(\mathbf{X},\mathbf{Y})$, define 
\begin{align}
    \Phi(Z) = 2\Mean\bracket*{g(X)\mathbbm{1}\cbrace*{Z \ge g(X)}}.
\end{align}
Then, we bound the excess error of $\hat{h}_k$ as
\begin{align}
    \Mean\bracket*{\mathcal{E}_Q(\hat{h}_k)} \le \Phi\paren*{\abs*{\hat\eta_k(X) - \eta(X)}}. \label{eq:excess-reg-err}
\end{align}
\cref{eq:excess-reg-err} indicates that a smaller approximation error results in a smaller excess error.

\paragraph{Implicit 1-NNs and implicit vicinity 1-NNs.}
Implicit 1-NNs, introduced by \textcite{gyorfi_distribution-free_2002}, are a crucial technique for analyzing the $k$-NN classifier. Given a point $X$ to be predicted, the implicit 1-NNs in the transfer learning setup are the 1-NNs of $X$ within $k$ disjoint batches consisting of subsamples of $(\mathbf{X},\mathbf{Y})_P$ and $(\mathbf{X},\mathbf{Y})_Q$ with sizes $\floor{\frac{n_P}{k}}$ and $\floor{\frac{n_Q}{k}}$, respectively. Let $B_1,...,B_k$ be the sets of the inputs appearing in the $i$th batch. The $i$th implicit 1-NN is defined as $X^*_{i} = \argmin_{X^* \in B_i}\rho(X^*,X)$ for $i=1,...,k$. Implicit 1-NNs behave similarly to $k$-NNs but are mutually independent, allowing, e.g., the use of concentration inequalities for independent random variables.

We also leverage the concept of implicit 1-NNs, but we employ the vicinity distance $\rho_{\mathcal{V}}$ instead of using the standard distance $\rho$, which we refer to as {\em implicit vicinity 1-NNs}. With the same definition of batches $B_1,...,B_k$, the implicit vicinity 1-NNs are defined as $\tilde{X}^i = \argmin_{X^* \in B_i}\rho_{\mathcal{V}}(X^*, X)$ for $i=1,...,k$.

\paragraph{Bounding via $\rho_{\mathcal{V}}$.}
We show an upper bound on the approximation error of $\hat\eta_k$ using the vicinity distance with the implicit vicinity 1-NNs, providing an upper bound on the excess error due to \cref{eq:excess-reg-err}.
\begin{theorem}\label{thm:reg-err-dist}
    Given $\alpha \in (0,1]$ and $\beta > 0$, suppose the target distribution $Q$ is $\mathrm{STN}(\alpha,\beta)$ with constants $C_\alpha$ and $C_\beta$. Then, there exist constant $C > 0$ and $c > 0$ (possibly) depending on $\alpha$ and $\beta$ such that for all $k > 1$ and for all $t > 0$, with probability, taken over the randomness of $(\mathbf{X},\mathbf{Y})$, at least $1-Ce^{-ckt^2}$,
    \begin{align}
        \abs*{\hat\eta_k(X) - \eta(X)} \le C_\alpha\Mean\bracket*{\rho^\alpha_{\dom{V}}\paren*{\tilde{X}^1, X}\middle|X} + \frac{1}{2}g(X) + t,
    \end{align}
    almost surely for the randomness of $X$.
\end{theorem}
\textcite{kpotufe_marginal_2021} provide a similar bound to \cref{thm:reg-err-dist}, but with the expected distance $C_\alpha\Mean\bracket{\rho^\alpha\paren{X^*_1, X}|X}$ instead of $C_\alpha\Mean\bracket{\rho^\alpha_{\dom{V}}\paren{\tilde{X}^*_1, X}|X} + \frac{1}{2}g(X)$. To achieve source sample-size consistency, the expected distance needs to vanish as the source sample size tends to infinity. However, under source and target distributions without support containment, the distance $\rho\paren{X^*_1, X}$ is larger than a non-zero positive constant. In contrast, the vicinity distance $\rho_{\mathcal{V}}\paren{\tilde{X}^*_1, X}$ can vanish as it takes the infimum over the vicinity set.

\subsection{Bounding via Dissimilarity Measure}
We now derive a high probability upper bound on the distance between the implicit (vicinity) 1-NN and the point to be predicted. To obtain this upper bound, we utilize a part of the analysis conducted by \textcite{pathak_new_2022}.
\begin{theorem}\label{thm:pathak-dist-sim}
    Given a distance $\rho$ over $\dom{X}$, define $\Delta(P,Q;r) = \int_{\dom{X}} \frac{1}{P_X(B(x,r))} Q(dx)$. Then, for $t > 0$,
    \begin{align}
        \Mean\bracket*{\mathbbm{1}\cbrace*{\min_{X^* \in B_1}\rho\paren*{X^*,X} > t}} \le \min\cbrace*{\frac{\Delta(P,Q;t)}{\floor{n_P/k}},\frac{\Delta(Q,Q;t)}{\floor{n_Q/k}}},
    \end{align}
    where the expectation is taken over the randomness of $(\mathbf{X},\mathbf{Y})$ and $X$.
\end{theorem}
By applying \cref{thm:pathak-dist-sim} with $\rho = \rho_{\dom{V}}$, we obtain a high probability upper bound on $\rho_{\dom{V}}\paren*{\tilde{X}^*_1, X}$ using our dissimilarity measure $\Delta_{\mathcal{V}}$. This result is essential for establishing the connection between the excess error of the $k$-NN classifier and our dissimilarity measure.

\subsection{Sketch Proof of \cref{thm:analyses-result}}
\cref{thm:analyses-result} is validated by combining \cref{thm:reg-err-dist}, \cref{thm:pathak-dist-sim}, and \cref{eq:excess-reg-err}. For simplicity, we only prove the case where $\tau,\psi > \alpha$ and left the other cases to \cref{sec:missing-proofs}. For a constant $\epsilon > 0$, define 
\begin{align}
    A(\epsilon,X) = \int_\epsilon^{D_{\dom{X}}} \Mean\bracket*{\mathbbm{1}\cbrace*{C_\alpha\rho^\alpha_{\dom{V}}\paren*{\tilde{X}^*_1, X} \ge t}|X} dt
\end{align}
Then, $C_\alpha\Mean\bracket*{\rho^\alpha_{\dom{V}}\paren{\tilde{X}^*_1, X}\middle|X} \le \epsilon + A(\epsilon,X)$. Hence, from \cref{eq:excess-reg-err} and \cref{thm:reg-err-dist}, there exists a random variable $\xi \ge 0$ depending on $(\mathbf{X},\mathbf{Y})$ and $X$ such that conditioned on $X$, $\xi \le t$ with probability at least $1-Ce^{-ckt^2}$, and
\begin{align}
    \Mean\bracket*{\mathcal{E}_Q(\hat{h}_k)} \le \Phi\paren*{2\paren*{\epsilon + A(\epsilon,X) + \xi}}. \label{eq:three-terms}
\end{align}
Under the assumptions of $\Delta_{\mathcal{V}}$-transfer- and $\Delta_{\mathcal{V}}$-self-exponents, applying \cref{thm:pathak-dist-sim} and adopting an approach similar to \textcite{kpotufe_marginal_2021}, we obtain that for some constant $C> 0$,
\begin{align}
    \Mean\bracket*{\mathcal{E}_Q(\hat{h}_k)} \le C\paren*{\epsilon^{1+\beta} + \min\cbrace*{\frac{\epsilon^{-\frac{\tau}{\alpha}+1}}{\floor{n_P/k}},\frac{\epsilon^{-\frac{\psi}{\alpha}+1}}{\floor{n_Q/k}}} + k^{-\frac{1+\beta}{2}}}, \label{eq:decomp-bound}
\end{align}
where the three terms in \cref{eq:decomp-bound} are bounds for the three terms in \cref{eq:three-terms}, respectively. To achieve the rate in \cref{thm:main}, we set $\epsilon = c\min\cbrace{\floor{\frac{n_P}{k}}^{-\frac{1}{\beta + \frac{\tau}{\alpha}}},\floor{\frac{n_Q}{k}}^{-\frac{1}{\beta + \frac{\psi}{\alpha}}}}$ for some constant $c > 0$ and assign $k$ as specified in the theorem statement.

\section{Experiment}

To confirm the tightness of \cref{thm:main} and the source sample-size consistency under the support non-containment environment, we carried out experiments on a synthetic dataset.

\paragraph{Data distribution.}
Let $\dom{X} = \RealSet$. For $\tau > 0$, $P_X$ has a density function proportional to $\paren{1-x^2}^{-\nicefrac{\tau}{2}}$ supported on $[-\frac{8^{\frac{1}{\alpha}}\cdot2-1}{8^{\frac{1}{\alpha}}\cdot2},\frac{8^{\frac{1}{\alpha}}\cdot2-1}{8^{\frac{1}{\alpha}}\cdot2}]$. $Q_X$ is the uniform distribution over $[-1,1]$, indicating that $\mathcal{X}_Q \not\subseteq \mathcal{X}_P$. Given $\alpha > 0$, the regression function $\eta$ is $\eta(x) = \frac{1}{2} + \frac{1}{2}\sgn(x)\abs{x}^\alpha$. With this setup, $Q$ is $\mathrm{STN}(\alpha,\beta)$ with $\beta = 1/\alpha$. The self-exponents are equivalent, i.e., $\psi_{\Delta_{\mathcal{V}}} = \psi_{\Delta_{\mathrm{PMW}}} = 1$. Due to the support non-containment, the $\Delta_{\mathrm{PMW}}$-transfer-exponent is $\infty$. On the other hand, the $\Delta_{\mathcal{V}}$-transfer-exponent is $\tau$.

\paragraph{Setup.}
We investigated the relationship between the source sample size $n_P$ and the excess error for $k$-NN classifiers using our parameter settings and those of \textcite{pathak_new_2022}. Note that the choices of $k$ differ between our method and that of \textcite{pathak_new_2022} due to differences between the $\Delta_{\mathrm{PMW}}$-transfer-exponent and the $\Delta_{\mathcal{V}}$-transfer-exponent. The training dataset was constructed by combining a sample from $P$ with size $n_P$ and a sample from $Q$ with size $n_Q$. We varied $n_P$ as $n_P \in \cbrace{2^8,2^9,...,2^{18}}$ while fixing $n_Q=10$. The test dataset, denoted as $(X'_1,Y'_1),...,(X'_m,Y'_m)$, was sampled from $Q$ with size $m=5000$. The empirical excess error was calculated using the following formula:
\begin{align}
    \mathcal{E}_{\mathrm{test},Q}(\hat{h}_k) = \frac{1}{m}\sum_{i=1}^m2g(X'_i)\mathbbm{1}\cbrace*{\hat{h}_k(X'_i) \ne h(X'_i)}.
\end{align}
We explored different parameter settings for $\alpha$ and $\tau$, with $\alpha \in \cbrace{\frac{1}{2}, \frac{1}{4}}$ and $\tau \in \cbrace{1, 2}$. For each parameter combination, we reported the average, first quartile, and third quartile of the excess error over 10 runs. All experiments were conducted on a machine equipped with an Intel Core i7-1065G7 CPU @ 1.30GHz, 16GB RAM. The implementation was done using Python 3.8.10 and the scikit-learn library (version 0.0.post11) for the $k$-NN classifier.

\begin{figure}[t]
    \centering
    \newcommand\scale{.35}
    \begin{subcaptionblock}{\scale\textwidth}
        \centering
        \includegraphics[width=\textwidth]{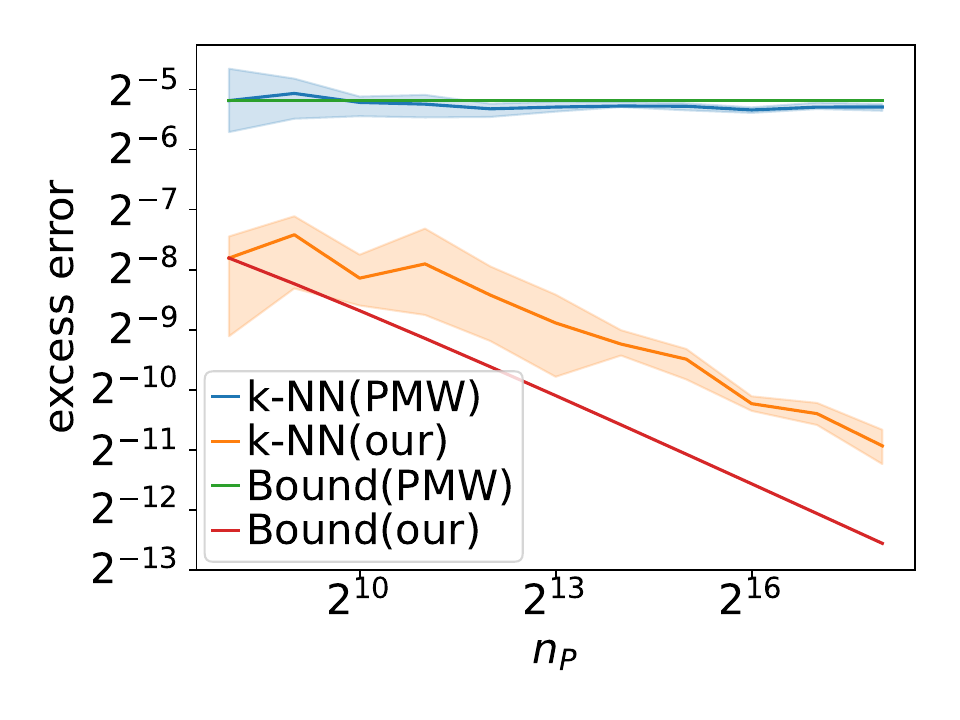}
        \caption{$\alpha=\frac{1}{2}, \tau = 1$}
    \end{subcaptionblock}
    \begin{subcaptionblock}{\scale\textwidth}
        \centering
        \includegraphics[width=\textwidth]{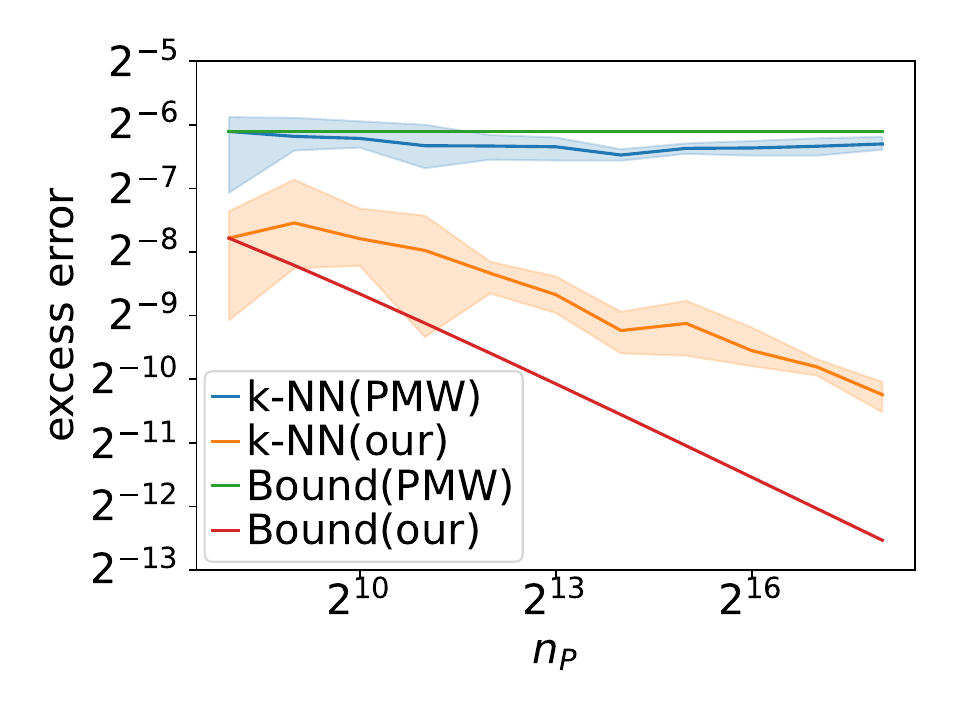}
        \caption{$\alpha=\frac{1}{4}, \tau = 1$}
    \end{subcaptionblock}
    \begin{subcaptionblock}{\scale\textwidth}
        \centering
        \includegraphics[width=\textwidth]{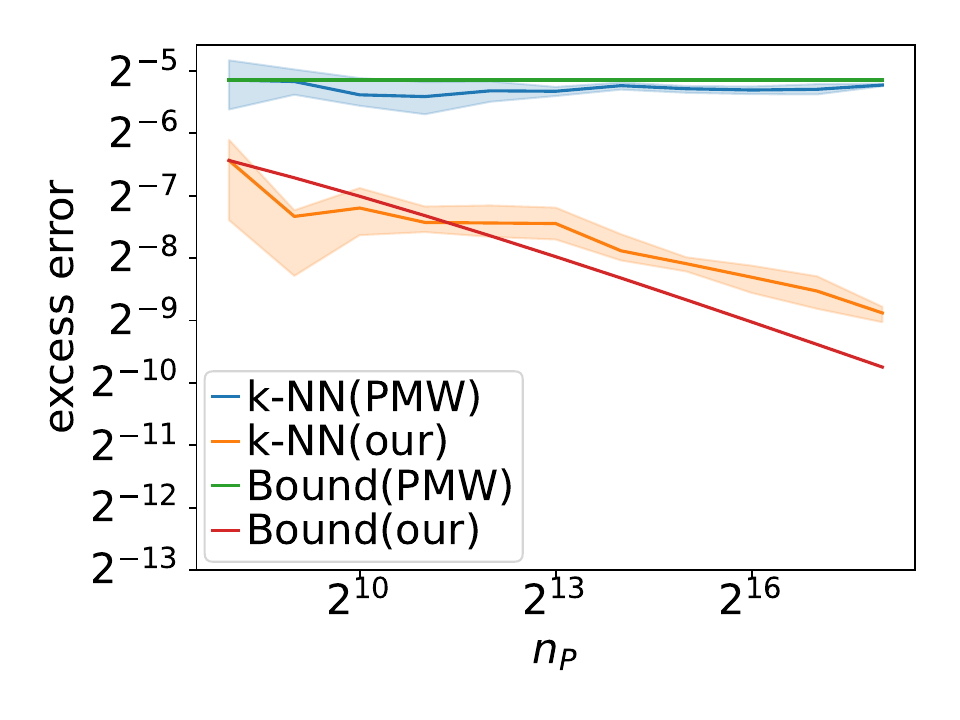}
        \caption{$\alpha=\frac{1}{2}, \tau = 2$}
    \end{subcaptionblock}
    \begin{subcaptionblock}{\scale\textwidth}
        \centering
        \includegraphics[width=\textwidth]{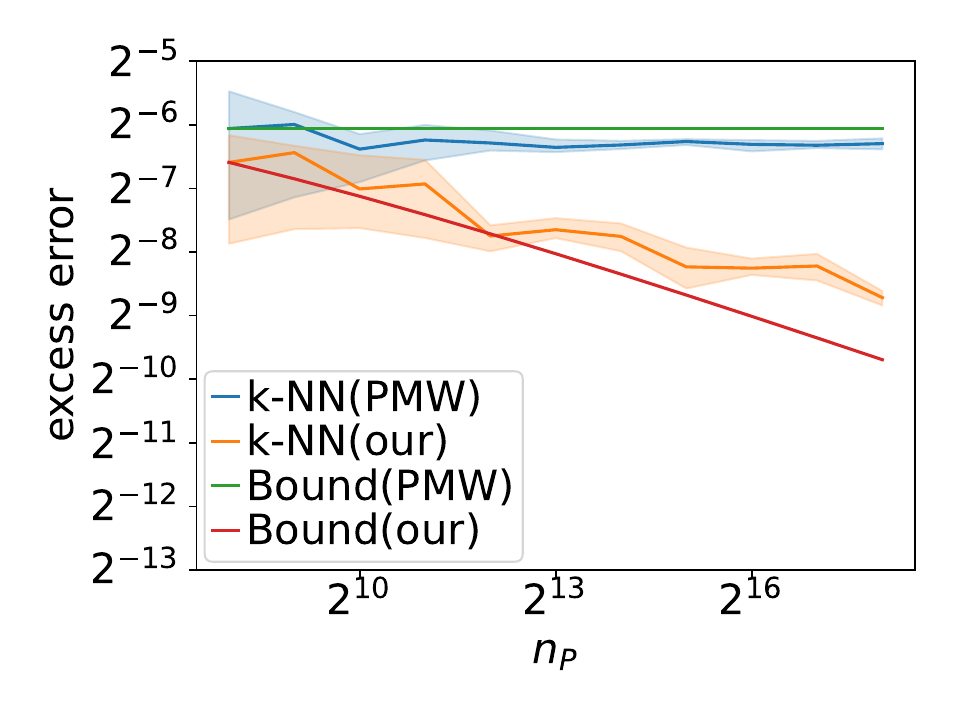}
        \caption{$\alpha=\frac{1}{4}, \tau = 2$}
    \end{subcaptionblock}
    \caption{Excess error $\mathcal{E}_Q$ vs. source sample size $n_P$. The horizontal axis represents the source sample size $n_P$, and the vertical axis represents the excess error. Both axes are plotted on a logarithmic scale, resulting in the slopes of the lines signifying the exponents of $n_P$ for each method. ''k-NN (our)`` and ''k-NN (PMW)`` denote the experimental results of $k$-NN classifiers with our and \textcite{pathak_new_2022}'s parameter settings, respectively. For guidance, we also plot the lines, ''Bound (our)`` and ''Bound (PMW)``, representing the upper bounds obtained by this paper and \textcite{pathak_new_2022}, respectively. }
    \label{fig:result}
\end{figure}

\paragraph{Results.}
\cref{fig:result} shows the log-log plots of the excess errors corresponding to the source sample sizes for each $\alpha$ and $\tau$. We also draw the upper bounds obtained by this paper and \textcite{pathak_new_2022} as guidlines. We adjusted the multiplicative constant of these guidelines so that the point at $n_P=2^8$ matches the experimental result of the corresponding $k$-NN at $n_P=2^8$. This adjustment helps to provide a clear visual comparison between the theoretical upper bounds and the empirical results.

For both our and \textcite{pathak_new_2022}'s $k$-NN, the slopes of the excess errors match the corresponding theoretical upper bounds for any parameter of $\alpha$ and $\tau$, demonstrating the tightness of the upper bounds across various situations. In \cref{fig:result}, the line representing \textcite{pathak_new_2022}'s $k$-NN does not decrease, indicating a failure to achieve source sample-size consistency. In contrast, our $k$-NN exhibits a decreasing excess error, signifying the successful achievement of source sample-size consistency.

\section{Related Work}\label{sec:related-work}
Theoretical works for covariate-shift typically provide upper bounds on the generalization error, which is the difference between the empirical average and expected losses. These works establish a connection between some divergence measure of the source and target distributions and the generalization error. For instance, \textcite{ben-david_theory_2010}'s analyses yield a generalization error bound that includes the $\dom{H}\Delta\dom{H}$-divergence, a measure of the discrepancy between the source and target distributions. Similarly, \textcite{park_calibrated_2020} introduce the {\em source-discrimination error}, which can be interpreted as a divergence between the source and target distributions, and provide a generalization error bound that incorporates this term. \textcite{aminian_information-theoretical_2022} employ the Kullback-Leibler~(KL) divergence between the source and target distributions to derive a generalization error bound under covariate-shift. Their techniques are applicable to a broader range of situations, as they do not make any assumptions about the source and target distributions. However, their approach may not be capable of confirming the consistency of the source sample size because their divergence measures remain positive even as the source sample size approaches infinity.

Several researchers leverage the likelihood ratio between the source and target distributions to derive upper bounds on the excess error under the covariate shift setup~\autocite{kpotufe_lipschitz_2017,ma_optimally_2023,feng_towards_2023}. In the same line of studies, the excess error analyses under the support non-containment setup were also provided~\autocite{liu_robust_2014,segovia-martin_double-weighting_2023,fang_generalizing_2023}. Their techniques can confirm the source sample-size consistency of their algorithms, but under the assumption that the learner has access to the likelihood ratio function. However, in real-world scenarios, the likelihood ratio function needs to be estimated using the training sample, which may introduce estimation errors. It is not certain that their methods exhibit the source sample-size consistency when employing the empirical estimation of the likelihood ratio.

Several techniques can confirm the existence of the source sample-size consistent algorithm~\autocite{pathak_new_2022,kpotufe_marginal_2021,galbraith_classification_2024}. We explored the comparison between our results and those obtained by \textcite{pathak_new_2022,kpotufe_marginal_2021} in \cref{sec:main} and demonstrated that our analysis always gives an upper bound with faster or competitive rates in \cref{prop:rel-exps}. \textcite{galbraith_classification_2024} introduced an ''average'' discrepancy to more tightly capture the behavior of the excess error under covariate-shift in classification. However, their technique does not account for the vicinity information and has the same limitations as the techniques by \textcite{pathak_new_2022,kpotufe_marginal_2021}, such as the inability to confirm the existence of the source sample-size consistent algorithm under support non-containment situations.

Many studies have explored the use of vicinity information to enhance prediction accuracy, beginning with the work of \textcite{chapelle_vicinal_2000}. The $k$-NN classifier can be interpreted as one such method, as it utilizes the $k$ nearest neighbors, which are points in the vicinity of a prediction target, to predict the label of the target point. Our approach differs in how the vicinity information is employed: while previous work primarily focuses on leveraging this information to improve prediction accuracy, we use it to refine our theoretical bounds.

\section{Conclusion}\label{sec:conclusion}
In this paper, we provide a novel analysis of excess error under the covariate-shift setup, demonstrating the usefulness of our new dissimilarity measure that utilizes vicinity information. Unlike existing analyses, our results can validate the consistency of the source sample size under certain situations where the support of the source sample does not contain that of the target distribution. We also demonstrate that our dissimilarity measure can provide faster rates than those provided by existing techniques, including \autocite{pathak_new_2022,kpotufe_marginal_2021}. Our findings contribute to bridging the gap between theoretical results and empirical observations in transfer learning, particularly in scenarios where the target and source distributions differ significantly. 

\paragraph{Border imacts and limitations}
There might be no additional societal impacts from those of the standard classification, as our focus is to leverage the multiple samples following different distributions to improve the classification accuracy. Our technique validate the existence of source sample-size consistent algorithm even in some support non-containment situations; however, it might fail to confirm the source-sample size consistency when the supports of the source and target distributions are significantly distant each other. Overcomming this limitation is our possible future direction.

\section*{Acknowledgement}
This work was partly supported by JSPS KAKENHI Grant Numbers JP23K13011, JST CREST JPMJCR21D3, and JSPS Grants-in-Aid for Scientific Research 23H00483.

\printbibliography

\appendix

\section{Missing Proofs}\label{sec:missing-proofs}

\subsection{Proof of \cref{prop:exist-bounds} and \cref{thm:analyses-result}}

With the choice of $k$ shown in the statement of \cref{thm:analyses-result}, there is a universal constant $c > 0$ such that for $n_P \ge c$ and $n_Q \ge c$, $n_P \ge 2k$ and $n_Q \ge 2k$. We can verify the upper bound in \cref{thm:main} holds when $n_P < c$ or $n_Q < c$ by adjusting the multiplicative constant, as the upper bound in \cref{thm:main} is decreasing in $n_P$ and $n_Q$. Therefore, we assume $n_P \ge 2k$ and $n_Q \ge 2k$ in the subsequent analyses.

The most parts of the proofs of \cref{prop:exist-bounds} and \cref{thm:analyses-result} are overlapped. As a non-overlapped part, we first demonstrate that in both cases of \cref{prop:exist-bounds} and \cref{thm:analyses-result}, we can validate that for a distance $\bar\rho$, which is either $\rho_{\mathcal{V}}$ or $\rho$, there exists a random variable $\xi \ge 0$ depending on $(\mathbf{X},\mathbf{Y})$ and $X$ such that conditioned on $X$, $\xi \le t$ with probability at least $1-e^{-ckt^2}$, and
\begin{align}
    \Mean\bracket*{\mathcal{E}_Q(\hat{h}_k)} \le \Mean\bracket*{2g(X)\mathbbm{1}\cbrace*{g(X) \le C\paren*{C_\alpha\Mean\bracket*{\min_{X^* \in B_i}\bar\rho^\alpha\paren*{X^*, X}\middle|X} + \xi}}}, \label{eq:universal-excess-bound}
\end{align}
where $C > 0$ is a universal constant.

To prove \cref{eq:universal-excess-bound} in the case of \cref{prop:exist-bounds}, we utilize the result by \textcite{kpotufe_marginal_2021}. \textcite{kpotufe_marginal_2021} reveal that the approximation error of $\hat\eta_k$ can be bounded above by the expected distance between an implicit 1-NN and $X$.
\begin{theorem}[\textcite{kpotufe_marginal_2021}]\label{thm:kpotufe-dist}
    Given $\alpha \in (0,1]$ and $\beta > 0$, suppose the target distribution $Q$ is $\mathrm{STN}(\alpha,\beta)$ with constants $C_\alpha$ and $C_\beta$. Then, there exist constants $C > 0$ and $c > 0$ (possibly) depending on $\alpha$ and $\beta$ such that with probability, taken over the randomness of $(\mathbf{X},\mathbf{Y})$, at least $1-Ce^{-ckt^2}$,
    \begin{align}
        \abs*{\hat\eta_k(X) - \eta(X)} \le C_\alpha\Mean\bracket*{\rho^\alpha\paren*{X^*_{1}, X}|X} + t, \label{eq:kpotufe-dist}
    \end{align}
    almost surely for the randomness of $X$.
\end{theorem}
The expected distance $\Mean\bracket{\rho^\alpha(X^*_1,X)|X}$ in \cref{eq:kpotufe-dist} corresponds to the bias incurred by the $k$-NN estimator $\hat\eta_k$.

From \cref{eq:excess-reg-err} and \cref{thm:reg-err-dist}, there exists a random variable $\xi \ge 0$ depending on $(\mathbf{X},\mathbf{Y})$ and $X$ such that conditioned on $X$, $\xi \le t$ with probability at least $1-e^{-ckt^2}$, and
\begin{align}
    \Mean\bracket*{\mathcal{E}_Q(\hat{h}_k)} \le \Mean\bracket*{2g(X)\mathbbm{1}\cbrace*{\frac{1}{2}g(X) \le C_\alpha\Mean\bracket*{\rho^\alpha_{\dom{V}}\paren*{\tilde{X}^*_1, X}\middle|X} + \xi}}.
\end{align}
Similarly, from \cref{eq:excess-reg-err} and \cref{thm:kpotufe-dist}, there exists a random variable $\xi \ge 0$ depending on $(\mathbf{X},\mathbf{Y})$ and $X$ such that conditioned on $X$, $\xi \le t$ with probability at least $1-e^{-ckt^2}$, and
\begin{align}
    \Mean\bracket*{\mathcal{E}_Q(\hat{h}_k)} \le \Mean\bracket*{2g(X)\mathbbm{1}\cbrace*{g(X) \le C_\alpha\Mean\bracket*{\rho^\alpha\paren*{X^*_1, X}\middle|X} + \xi}}.
\end{align}
Consequently, \cref{eq:universal-excess-bound} is verified in both cases.

\paragraph{Universal analyses for proving \cref{prop:exist-bounds} and \cref{thm:analyses-result}.}
For a constant $\epsilon > 0$, define 
\begin{align}
    A(\epsilon,X) = \int_\epsilon^{\infty} \Mean\bracket*{\mathbbm{1}\cbrace*{C_\alpha\min_{X^* \in B_i}\bar\rho^\alpha\paren*{X^*, X} \ge t} \middle| X} dt.
\end{align}
Then, we have 
\begin{align}
    \Mean\bracket*{C_\alpha\min_{X^* \in B_i}\bar\rho\paren*{X^*, X}\middle|X} = \int_0^\infty \Mean\bracket*{\mathbbm{1}\cbrace*{C_\alpha\min_{X^* \in B_i}\bar\rho^\alpha\paren*{X^*, X} > t}\middle|X} dt 
    \le \epsilon + A(\epsilon, X).
\end{align}
Hence,
\begin{align}
    \Mean\bracket*{\mathcal{E}_Q(\hat{h}_k)} \le \Mean\bracket*{2g(X)\mathbbm{1}\cbrace*{g(X) \le C\paren*{\epsilon + A(\epsilon, X) + \xi}}}. \label{eq:before-decomp}
\end{align}

For any positive reals $a$ and $b_1,...,b_m$, we have
\begin{align}
    \mathbbm{1}\cbrace*{a \le \sum_{i=1}^mb_i} \le \mathbbm{1}\cbrace*{a \le m\max_{i=1,...,m}b_i} 
     \le \sum_{i=1}^m\mathbbm{1}\cbrace*{a \le mb_i}. \label{eq:real-decomp}
\end{align}
Applying \cref{eq:real-decomp} to \cref{eq:before-decomp} yields
\begin{multline}
    \Mean\bracket*{\mathcal{E}_Q(\hat{h}_k)} \le \Mean\bracket*{2g(X)\mathbbm{1}\cbrace*{g(X) \le 3C\epsilon}}\\+ \Mean\bracket*{2g(X)\mathbbm{1}\cbrace*{g(X) \le 3C\xi}} + \Mean\bracket*{2g(X)\mathbbm{1}\cbrace*{g(X) \le 3CA(\epsilon, X)}}. \label{eq:decompose}
\end{multline}
We will provide upper bounds on the three terms in \cref{eq:decompose}.

\paragraph{First term in \cref{eq:decompose}.}
From \cref{def:noise}, we have
\begin{align}
    \Mean\bracket*{2g(X)\mathbbm{1}\cbrace*{g(X) \le 3C\epsilon}} \le 6C\epsilon Q_X\paren*{g(X) \le 3C\epsilon} \le 2C_\beta\paren*{3C\epsilon}^{1+\beta}. \label{eq:first-bound}
\end{align}

\paragraph{Second term in \cref{eq:decompose}.}
We utilize Lemma 4 of \textcite{kpotufe_marginal_2021}. 
\begin{lemma}[\textcite{kpotufe_marginal_2021}]\label{lem:kpotufe-4}
    Let $Z$ be a random variable depending on $(\mathbf{X},\mathbf{Y})$ and $X$ such that for $t > 0$,
    \begin{align}
        \Mean\bracket*{\mathbbm{1}\cbrace*{Z \ge t}} \le Ce^{-ckt^2},
    \end{align}
    for some constants $C > 0$ and $c > 0$. Then, we have
    \begin{align}
        \Mean\bracket*{g(X)\mathbbm{1}\cbrace*{g(X) \le Z}} \le 3CC_\beta\paren*{\frac{1+\beta}{ck}}^{\frac{1+\beta}{2}}.
    \end{align}
\end{lemma}
Applying \cref{lem:kpotufe-4} yields 
\begin{align}
    \Mean\bracket*{g(X)\mathbbm{1}\cbrace*{g(X) \le 3C\xi}} \le Ck^{-\frac{1+\beta}{2}}, \label{eq:second-bound}
\end{align}
for some constant $C > 0$.

\paragraph{Third term in \cref{eq:decompose}.}
Let $\bar{D}_{\dom{X}}$ be the diameter of $\dom{X}$ with respect to $\bar\rho$. Applying \cref{thm:pathak-dist-sim} to the third term in \cref{eq:decompose} yields
\begin{multline}
    \Mean\bracket*{g(X)\mathbbm{1}\cbrace*{g(X) \le 3CA(\epsilon, X)}} \le 3C\int_\epsilon^{\bar{D}_{\dom{X}}} \min\cbrace*{\frac{\Delta_{\mathcal{V}}(P,Q;(t/C_\alpha)^{\nicefrac{1}{\alpha}})}{\floor{n_P/k}},\frac{\Delta_{\mathcal{V}}(Q,Q;(t/C_\alpha)^{\nicefrac{1}{\alpha}})}{\floor{n_Q/k}}} dt. \label{eq:bound-by-dist}
\end{multline}
Under the assumptions of $\Delta_{\mathcal{V}}$-transfer-exponent and $\Delta_{\mathcal{V}}$-self-exponent, \cref{eq:bound-by-dist} is bounded above by
\begin{align}
    C\int_\epsilon^{\bar{D}_{\dom{X}}}\min\cbrace*{\frac{t^{-\frac{\tau}{\alpha}}}{\floor{n_P/k}},\frac{t^{-\frac{\psi}{\alpha}}}{\floor{n_Q/k}}}dt.
\end{align}
We analyze the integration of $t^{-\gamma}$ for $\gamma > 0$, as exchanging the minimum and integral will give an upper bound. Some elementary calculations give
\begin{align}
    \int_\epsilon^{\bar{D}_{\dom{X}}} t^{-\gamma} dt = \begin{dcases}
      \frac{1}{1-\gamma}\paren*{\bar{D}_{\dom{X}}^{1-\gamma} - \epsilon^{1-\gamma}}  & \textif \gamma \ne 1, \\
      \ln\paren*{\frac{\bar{D}_{\dom{X}}}{\epsilon}}  & \textif \gamma = 1.
    \end{dcases}
\end{align}
Hence,
\begin{align}
    \int_\epsilon^{\bar{D}_{\dom{X}}} t^{-\gamma} dt \le \begin{dcases}
      C\paren*{\frac{\bar{D}_{\dom{X}}}{\epsilon}}^{\gamma-1}  & \textif \gamma > 1, \\
      \log\paren*{\frac{\bar{D}_{\dom{X}}}{\epsilon}}  & \textif \gamma = 1, \\
      C  & \textif \gamma < 1.
    \end{dcases}
\end{align}
for some constant $C > 0$. Consequentially, letting 
\begin{align}
    u_\gamma(x) = \begin{dcases}
        x^{-(\gamma - 1)} & \textif \gamma > 1, \\
        \log(1/x) & \textif \gamma = 1, \\
        1 & \textif \gamma < 1,
    \end{dcases},
\end{align}
for $x > 0$, \cref{eq:bound-by-dist} is bounded above by
\begin{align}
    C\min\cbrace*{\frac{u_{\frac{\tau}{\alpha}}\paren*{\frac{\epsilon}{\bar{D}_{\dom{X}}}}}{\floor{n_P/k}},\frac{u_{\frac{\psi}{\alpha}}\paren*{\frac{\epsilon}{\bar{D}_{\dom{X}}}}}{\floor{n_Q/k}}}. \label{eq:third-bound}
\end{align}

\paragraph{Rest of the proof.}
By combining \cref{eq:first-bound,eq:second-bound,eq:third-bound}, we have
\begin{align}
    \Mean\bracket*{\mathcal{E}_Q(\hat{h}_k)} \le C\paren*{\epsilon^{1+\beta} + k^{-\frac{1+\beta}{2}} + \min\cbrace*{\frac{u_{\frac{\tau}{\alpha}}\paren*{\frac{\epsilon}{\bar{D}_{\dom{X}}}}}{\floor{n_P/k}},\frac{u_{\frac{\psi}{\alpha}}\paren*{\frac{\epsilon}{\bar{D}_{\dom{X}}}}}{\floor{n_Q/k}}}}.
\end{align}

We can obtain the desired rate in \cref{thm:main} by setting
\begin{align}
    \frac{\epsilon}{\bar{D}_{\dom{X}}} = c\min\cbrace*{\floor*{\frac{n_P}{k}}^{-\frac{1}{\beta + \max\cbrace*{1,\frac{\tau}{\alpha}}}},\floor*{\frac{n_Q}{k}}^{-\frac{1}{\beta + \max\cbrace*{1,\frac{\psi}{\alpha}}}}},
\end{align}
for some constant $c > 0$ so that $\epsilon \le \bar{D}_{\dom{X}}$ and assigning $k$ as shown in the statement. Note that with $k$ shown in the statement, we have
\begin{align}
    \max\cbrace*{n_P^{\frac{2}{2 + \beta + \max\cbrace{1,\frac{\tau}{\alpha}}}}, n_Q^{\frac{2}{2 + \beta + \max\cbrace{1,\frac{\psi}{\alpha}}}}} \le k \le 2\max\cbrace*{n_P^{\frac{2}{2 + \beta + \max\cbrace{1,\frac{\tau}{\alpha}}}}, n_Q^{\frac{2}{2 + \beta + \max\cbrace{1,\frac{\psi}{\alpha}}}}}. \label{eq:k-lower-upper}
\end{align}
Assume $\tau \ne \alpha$ and $\psi \ne \alpha$. Then, assigning $\epsilon$ yields
\begin{multline}
    \Mean\bracket*{\mathcal{E}_Q(\hat{h}_k)} \le C\paren[\Bigg]{\min\cbrace*{\floor*{\frac{n_P}{k}}^{-\frac{1+\beta}{\beta + \max\cbrace*{1,\frac{\tau}{\alpha}}}},\floor*{\frac{n_Q}{k}}^{-\frac{1+\beta}{\beta + \max\cbrace*{1,\frac{\psi}{\alpha}}}}} + k^{-\frac{1+\beta}{2}} \\ + \min\cbrace*{\floor*{\frac{n_P}{k}}^{-\frac{1 + \beta}{\beta + \max\cbrace*{1,\frac{\tau}{\alpha}}}},\floor*{\frac{n_Q}{k}}^{-\frac{1+\beta}{\beta + \max\cbrace*{1,\frac{\psi}{\alpha}}}}}}, \label{eq:bound-with-k}
\end{multline}
where we use $\min\cbrace{\min\cbrace{a^{\alpha_1},b^{\beta_1}}/a^{\alpha_2},\min\cbrace{a^{\alpha_1},b^{\beta_1}}/b^{\beta_2}} \le \min\cbrace{a^{\alpha_1-\alpha_2},b^{\beta_1-\beta_2}}$ for $a, b, \alpha_1, \alpha_2, \beta_1, \beta_2 > 0$ to obtain the third term. From \cref{eq:k-lower-upper} and the assumption of $n_P \ge 2k$ and $n_Q \ge 2k$, we have
\begin{align}
    \floor*{\frac{n_P}{k}} \ge& cn_P^{\frac{\beta + \max\cbrace{1,\frac{\tau}{\alpha}}}{2+\beta+\max\cbrace{1,\frac{\tau}{\alpha}}}} \\
    \floor*{\frac{n_Q}{k}} \ge& cn_Q^{\frac{\beta + \max\cbrace{1,\frac{\psi}{\alpha}}}{2+\beta+\max\cbrace{1,\frac{\psi}{\alpha}}}},
\end{align}
for some constant $c > 0$. Substituting this and \cref{eq:k-lower-upper} into \cref{eq:bound-with-k} yields the claim.

\subsection{Proof of \cref{prop:rel-exps}}
\begin{proof}[Proof of \cref{prop:rel-exps}]
    By definitions, for all $P$, $Q$, and $r > 0$,
    \begin{align}
        \Delta_{\mathcal{V}}(P,Q;r) \le \Delta_{\mathrm{PMW}}(P,Q;r),
    \end{align}
    which verifies the statement about the relationship between $\Delta_{\mathcal{V}}$ and $\Delta_{\mathrm{PMW}}$. Also, by definitions, for all $Q$ and $r > 0$,
    \begin{align}
        \Delta_{\mathrm{PMW}}(Q,Q;r) \le \Delta_{\mathrm{DM}}(Q,Q;r),
    \end{align}
    by which we can verify the relationship between $\psi_{\Delta_{\mathrm{PMW}}}$ and $\psi_{\Delta_{\mathrm{DM}}}$.
    
    Consider a cover of $\dom{X}_Q$ with $\mathcal{N}(\dom{X}_Q,\rho,r)$ balls of radius $\frac{r}{2}$ whose centers are $x_1,...,x_{\mathcal{N}(\dom{X}_Q,\rho,\frac{r}{2})}$. Then, we have
    \begin{align}
        \Delta_{\mathrm{PMW}}(Q,Q;r) \le& \sum_i \int_{B(x_i,\frac{r}{2})} \frac{1}{Q_X(B(x,r))} Q_X(dx) \\
        \le& \sum_i \int_{B(x_i,\frac{r}{2})} \frac{1}{Q_X(B(x_i,\frac{r}{2}))} Q_X(dx) \\
        \le& \mathcal{N}\paren*{\dom{X}_Q,\rho,\frac{r}{2}},
    \end{align}
    which yields $\psi_{\Delta_{\mathrm{PMW}}} \le \psi_{\Delta_{\mathrm{BCN}}}$.
    
    Lastly, we prove the inequality $\tau_{\Delta_{\mathrm{PMW}}} \le \tau_{\Delta_{\mathrm{KM}}} + \min\cbrace{\psi_{\Delta_{\mathrm{DM}}},\psi_{\Delta_{\mathrm{BCN}}}}$. Suppose $\Delta_{\mathrm{KM}}(P,Q;r) \le Cr^{-\tau}$. Then, we have
    \begin{align}
        \Delta_{\mathrm{PMW}}(Q,Q;r) =& \int \frac{1}{P_X(B(x,r))} Q_X(dx) \\
        \le& Cr^{-\tau}\int \frac{1}{Q_X(B(x,r))} Q_X(dx) = Cr^{-\tau}\Delta_{\mathrm{PMW}}(Q,Q;r),
    \end{align}
    which gives the desired inequality.
\end{proof}

\subsection{Proof of \cref{thm:reg-err-dist}}
\begin{proof}[Proof of \cref{thm:reg-err-dist}]
    From the $\alpha$-H\"older continuity assumption, we have, conditioned on $X$,
    \begin{align}
        \abs*{\hat\eta(X) - \eta(X)} =& \abs*{\frac{1}{k}\sum_{i=1}^kY_{(i)} - \eta(X)} \\
        \le& \abs*{\frac{1}{k}\sum_{i=1}^k\paren*{Y_{(i)} - \eta(X_{(i)})}} + \abs*{\frac{1}{k}\sum_{i=1}^k\eta(X_{(i)}) - \eta(X)} \\
        \le& \abs*{\frac{1}{k}\sum_{i=1}^k\paren*{Y_{(i)} - \eta(X_{(i)})}} + \frac{C_\alpha}{k}\sum_{i=1}^k\rho^\alpha\paren*{X_{(i)}, X}. \label{eq:reg-err-decomp}
    \end{align}
    
    Applying the Hoeffding inequality into the first term in \cref{eq:reg-err-decomp} with conditioned on $X_1,...,X_{n_P+n_Q}$ yields that the first term in \cref{eq:reg-err-decomp} is bounded above by $\frac{t}{2}$ with probability at least $1 - 2e^{-kt^2/2}$.
    
    Let us focus on the second term in \cref{eq:reg-err-decomp}. For any distinct indices $j_1,...,j_k \in \cbrace{1,...,n_P+n_Q}$, we have
    \begin{align}
        \frac{1}{k}\sum_{i=1}^k\rho^\alpha\paren*{X_{(i)}, X} \le \frac{1}{k}\sum_{i=1}^k\rho^\alpha\paren*{X_{j_i}, X},
    \end{align}
    as $X_{(1)},...,X_{(k)}$ are the $k$-NNs of $X$. We set these indices as the indices of $k$-NNs in terms of the vicinity distance $\rho_{\mathcal{V}}$. Letting $\tilde{X}_{(1)},...,\tilde{X}_{(k)}$ be such $k$-NNs, we have
    \begin{align}
        \frac{1}{k}\sum_{i=1}^k\rho^\alpha\paren*{X_{(i)}, X} \le \frac{1}{k}\sum_{i=1}^k\rho^\alpha\paren*{\tilde{X}_{(i)}, X}. \label{eq:dist-vicinity-knn}
    \end{align}
    
    In the same manner, \cref{eq:dist-vicinity-knn} is bounded above by the average distance of the implicit vicinity 1-NNs, i.e.,
    \begin{align}
        \frac{1}{k}\sum_{i=1}^k\rho^\alpha\paren*{\tilde{X}_{(i)}, X} \le \frac{1}{k}\sum_{i=1}^k\rho^\alpha\paren*{\tilde{X}^*_{i}, X}.
    \end{align}
    
    From the triangle inequality and the definition of the vicinity set in \cref{eq:vicinity}, for any points $X'_i \in \mathcal{V}(X)$, we have
    \begin{align}
        \frac{1}{k}\sum_{i=1}^k\rho^\alpha\paren*{\tilde{X}^*_{i}, X} \le& \frac{1}{k}\sum_{i=1}^k\paren*{\rho^\alpha\paren*{\tilde{X}^*_{i}, X'_i} + \rho\paren*{X'_i,X}} \\
        \le& \frac{1}{k}\sum_{i=1}^k\paren*{\rho^\alpha\paren*{\tilde{X}^*_{i}, X'_i} + \frac{1}{2}g(X)}.
    \end{align}
    
    By the definition of the infimum, for any $\epsilon > 0$, there exist $X'_i$ such that
    \begin{align}
        \frac{1}{k}\sum_{i=1}^k\paren*{\rho^\alpha\paren*{\tilde{X}^*_{i}, X'_i} + \frac{1}{2}g(X)} \le& \frac{1}{k}\sum_{i=1}^k\rho^\alpha_{\mathcal{V}}\paren*{\tilde{X}^*_{i}, X} + \epsilon + \frac{1}{2}g(X).
    \end{align}
    
    From the arbitrariness of $\epsilon > 0$, we have
    \begin{align}
        \frac{1}{k}\sum_{i=1}^k\rho^\alpha\paren*{\tilde{X}^*_{i}, X} \le& \frac{1}{k}\sum_{i=1}^k\rho^\alpha_{\mathcal{V}}\paren*{\tilde{X}^*_{i}, X} + \frac{1}{2}g(X). \label{eq:dist-vicinity-1nn}
    \end{align}
    
    We now apply the Hoeffding inequality into the first term of \cref{eq:dist-vicinity-1nn}. Then, the first term of \cref{eq:dist-vicinity-1nn} is bounded above as
    \begin{align}
        \frac{C_\alpha}{k}\sum_{i=1}^k\rho^\alpha_{\mathcal{V}}\paren*{\tilde{X}^*_{i}, X} \le C_\alpha\Mean\bracket*{\rho^\alpha\paren*{X^*_1, X} \middle| X} + \frac{t}{2}, \label{eq:dist-exp-vicinity}
    \end{align}
    with probability at least $1- 2e^{-kt^2/2C_\alpha D_{\dom{X}^\alpha}}$. The claim is verified by combining \cref{eq:reg-err-decomp,eq:dist-vicinity-1nn,eq:dist-exp-vicinity} and the union bound.
\end{proof}

\subsection{Proof of \cref{thm:pathak-dist-sim}}
\begin{proof}[Proof of \cref{thm:pathak-dist-sim}]
    Remark that $B_1$ contains the subsamples from $(\mathbf{X},\mathbf{Y})_P$ with the size $\floor{n_P/k}$ and $(\mathbf{X},\mathbf{Y})_Q$ with the size $\floor{n_Q/k}$. By the mutual independence among $X_1,...,X_{n_P+n_Q}$, we have
    \begin{align}
        & \Mean\bracket*{\mathbbm{1}\cbrace*{\min_{X^* \in B_1}\rho(X^*,X) > t}|X} \\
        =& \Mean\bracket*{\mathbbm{1}\cbrace*{\forall X^* \in B_1, \rho(X^*,X) > t}|X} \\
        =& \prod_{X^* \in B_1}\Mean\bracket*{\mathbbm{1}\cbrace*{\rho(X^*,X) > t}|X} \\
        =& \paren*{1 - P_X\paren*{B(X,t)}}^{\floor*{\frac{n_P}{k}}}\paren*{1 - Q_X\paren*{B(X,t)}}^{\floor*{\frac{n_Q}{k}}} \\
        \le& \paren*{\floor*{\frac{n_P}{k}}P_X\paren*{B(X,t)} + \floor*{\frac{n_Q}{k}}Q_X\paren*{B(X,t)}}^{-1},
    \end{align}
    where the last inequality follows from $(1-p)^n(1-q)^m \le \exp\paren{-(np+mq)} \le \paren{np + mq}^{-1}$. Taking the expectation over $X$ yields
    \begin{align}
        &\Mean\bracket*{\mathbbm{1}\cbrace*{\min_{X^* \in B_1}\rho(X^*,X) > t}} \\
        \le& \Mean\bracket*{\paren*{\floor*{\frac{n_P}{k}}P_X\paren*{B(X,t)} + \floor*{\frac{n_Q}{k}}Q_X\paren*{B(X,t)}}^{-1}} \\
        \le& \Mean\bracket*{\min\cbrace*{\frac{1}{\floor*{n_P/k}P_X\paren*{B(X,t)}},\frac{1}{\floor*{n_Q/k}Q_X\paren*{B(X,t)}}}} \\
        \le& \min\cbrace*{\Mean\bracket*{\frac{1}{\floor*{n_P/k}P_X\paren*{B(X,t)}}},\Mean\bracket*{\frac{1}{\floor*{n_Q/k}Q_X\paren*{B(X,t)}}}},
    \end{align}
    which concludes the claim.
\end{proof}

\end{document}